\documentclass[11pt]{article}
\pdfoutput=1
\usepackage[
  margin=0.83in,
  headheight=12pt,
  headsep=25pt,
  includefoot,
  footskip=30pt,
]{geometry}
\usepackage{tikz}
\usepackage{graphicx}
\usepackage{comment}
\usepackage{listings}
\usepackage{xcolor}
\usepackage{inconsolata}
\usepackage{makecell}
\usepackage{tabularx} 
\usepackage{adjustbox}
\usepackage{rotating}

\definecolor{commentcolour}{rgb}{0.3,0.7,0.2}
\definecolor{backcolour}{rgb}{0.98,0.98,0.98}

\lstdefinelanguage{markdown}{
    comment=[l]{\#},
    morestring=[s]{```}{```},
    commentstyle=\color{commentcolour}\bfseries,
    stringstyle=\color{blue},
    basicstyle=\scriptsize\ttfamily,
    showstringspaces=false,
    breaklines=true,
    breakautoindent=false,
    breakindent=0pt,
    backgroundcolor=\color{backcolour},
}
\lstdefinestyle{mystyle}{
    morekeywords={self},
    basicstyle=\scriptsize\ttfamily,
    keywordstyle=\color{blue},
    commentstyle=\color{commentcolour}\bfseries,
    breaklines=true,
    breakautoindent=false,
    showstringspaces=false,
    backgroundcolor=\color{backcolour},
    stringstyle=\color{red},
}
\lstdefinelanguage{PythonPlus}[]{Python}{
  alsoother={@},
  morekeywords=[1]{,as,assert,nonlocal,with,yield,self,True,False,None} % Python builtin
  morekeywords=[2]{,__init__,__add__,__mul__,__div__,__sub__,__call__,__getitem__,__setitem__,__eq__,__ne__,__nonzero__,__rmul__,__radd__,__repr__,__str__,__get__,__truediv__,__pow__,__name__,__future__,__all__,}, % magic methods
  morekeywords=[3]{,object,type,isinstance,copy,deepcopy,zip,enumerate,reversed,list,set,len,dict,tuple,range,xrange,append,execfile,real,imag,reduce,str,repr,}, % common functions
  morekeywords=[4]{,Exception,NameError,IndexError,SyntaxError,TypeError,ValueError,OverflowError,ZeroDivisionError,}, % errors
  morekeywords=[5]{,ode,fsolve,sqrt,exp,sin,cos,arctan,arctan2,arccos,pi, array,norm,solve,dot,arange,isscalar,max,sum,flatten,shape,reshape,find,any,all,abs,plot,linspace,legend,quad,polyval,polyfit,hstack,concatenate,vstack,column_stack,empty,zeros,ones,rand,vander,grid,pcolor,eig,eigs,eigvals,svd,qr,tan,det,logspace,roll,min,mean,cumsum,cumprod,diff,vectorize,lstsq,cla,eye,xlabel,ylabel,squeeze,}, % numpy / math
}

\usepackage{tikz}
\usepackage{amsmath} % for matrices
\usepackage{array} % for tabular alignment
\usetikzlibrary{matrix, arrows}
\usepackage{amsmath,amssymb}
\usepackage{amsthm}
\usepackage{mathtools}
\usepackage{xspace}
\usepackage[noend]{algorithmic}
\usepackage[ruled,vlined]{algorithm2e}
\usepackage{url}
\usepackage{makeidx}
\usepackage{enumerate}
\usepackage{epstopdf}
\usepackage{booktabs}
\usepackage{color}
\usepackage[utf8]{inputenc}
\usepackage{thm-restate}
\usepackage{scalerel,stackengine}
\usepackage[shortlabels]{enumitem}
\usepackage{xr}
\usepackage{fancyvrb}
\usepackage{xcolor}
\usepackage{bold-extra}
\usepackage{arydshln}
\usepackage[small]{caption}
\usepackage{subcaption}
\usepackage[most]{tcolorbox}
\usepackage{fvextra}
\usepackage{float}
\usepackage{alltt}
\usepackage{soul}
\usepackage{MnSymbol,wasysym}
\usepackage{fancyvrb}
\usepackage{multirow}
\usepackage[final]{hyperref}
\usepackage[bottom]{footmisc}
\usepackage{diagbox}
\usepackage{mdframed}
\usepackage{todonotes}
\setuptodonotes{inline}
\usepackage{tikz}
\usetikzlibrary{shapes,calc,positioning}

\global

\tcbset{
  aibox/.style={
    width=\textwidth,
    top=0pt, bottom=0pt, left=5pt, right=5pt,
    colback=white,
    colframe=black,
    colbacktitle=black,
    enhanced,
    center,
    attach boxed title to top left={yshift=-0.1in,xshift=0.15in},
    boxed title style={boxrule=0pt,colframe=white,},
  }
}
\newtcolorbox{AIbox}[2][]{aibox,title=#2,#1}

\definecolor{aigold}{RGB}{244,210, 1} 
\definecolor{aigreen}{RGB}{210,244,211} 

\sethlcolor{aigreen}

\definecolor{aired}{RGB}{255,180,181}

\newcommand{\phivision}{Phi-3.5-Vision\xspace}

\newcommand{\datasetcell}[3]{\makecell{ \large #1  \\  \tiny (#2) \tiny #3   }  }

\newtcbox{\mybox}[1][green]{on line,
arc=0pt,outer arc=0pt,colback=#1!10!white,colframe=#1!50!black,
boxsep=0pt,left=0pt,right=0pt,top=0pt,bottom=0pt,
boxrule=0pt,bottomrule=0pt,toprule=0pt}

\begin{document}

\title{Phi-3 Technical Report: \\
A Highly Capable Language Model Locally on Your Phone}

\author{Microsoft}
\date{}

\maketitle

\begin{abstract}
We introduce \textbf{phi-3-mini}, a 3.8 billion parameter language model trained on 3.3 trillion tokens, whose overall performance, as measured by both academic benchmarks and internal testing, rivals that of models such as Mixtral 8x7B and GPT-3.5 (e.g., \textbf{phi-3-mini} achieves 69\% on MMLU and 8.38 on MT-bench), despite being small enough to be deployed on a phone. Our training dataset is a scaled-up version of the one used for \textbf{phi-2}, composed of heavily filtered publicly available web data and synthetic data. The model is also further aligned for robustness, safety, and chat format.
We also provide parameter-scaling results with a 7B, 14B models trained for 4.8T tokens, called \textbf{phi-3-small}, \textbf{phi-3-medium}, both significantly more capable than \textbf{phi-3-mini} (e.g., respectively 75\%, 78\% on MMLU, and 8.7, 8.9 on MT-bench).
To enhance multilingual, multimodal, and long-context capabilities, we introduce three models in the \textbf{phi-3.5} series: \textbf{phi-3.5-mini}, \textbf{phi-3.5-MoE}, and \textbf{phi-3.5-Vision}. The \textbf{phi-3.5-MoE}, a 16 x 3.8B MoE model with 6.6 billion active parameters, achieves superior performance in language reasoning, math, and code tasks compared to other open-source models of similar scale, such as Llama 3.1 and the Mixtral series, and on par with Gemini-1.5-Flash and GPT-4o-mini. Meanwhile, \textbf{phi-3.5-Vision}, a 4.2 billion parameter model derived from \textbf{phi-3.5-mini}, excels in reasoning tasks and is adept at handling both single-image and text prompts, as well as multi-image and text prompts.

\end{abstract}

\section{Introduction}
\begin{figure}
    \centering
    \includegraphics[width=0.30\textwidth]{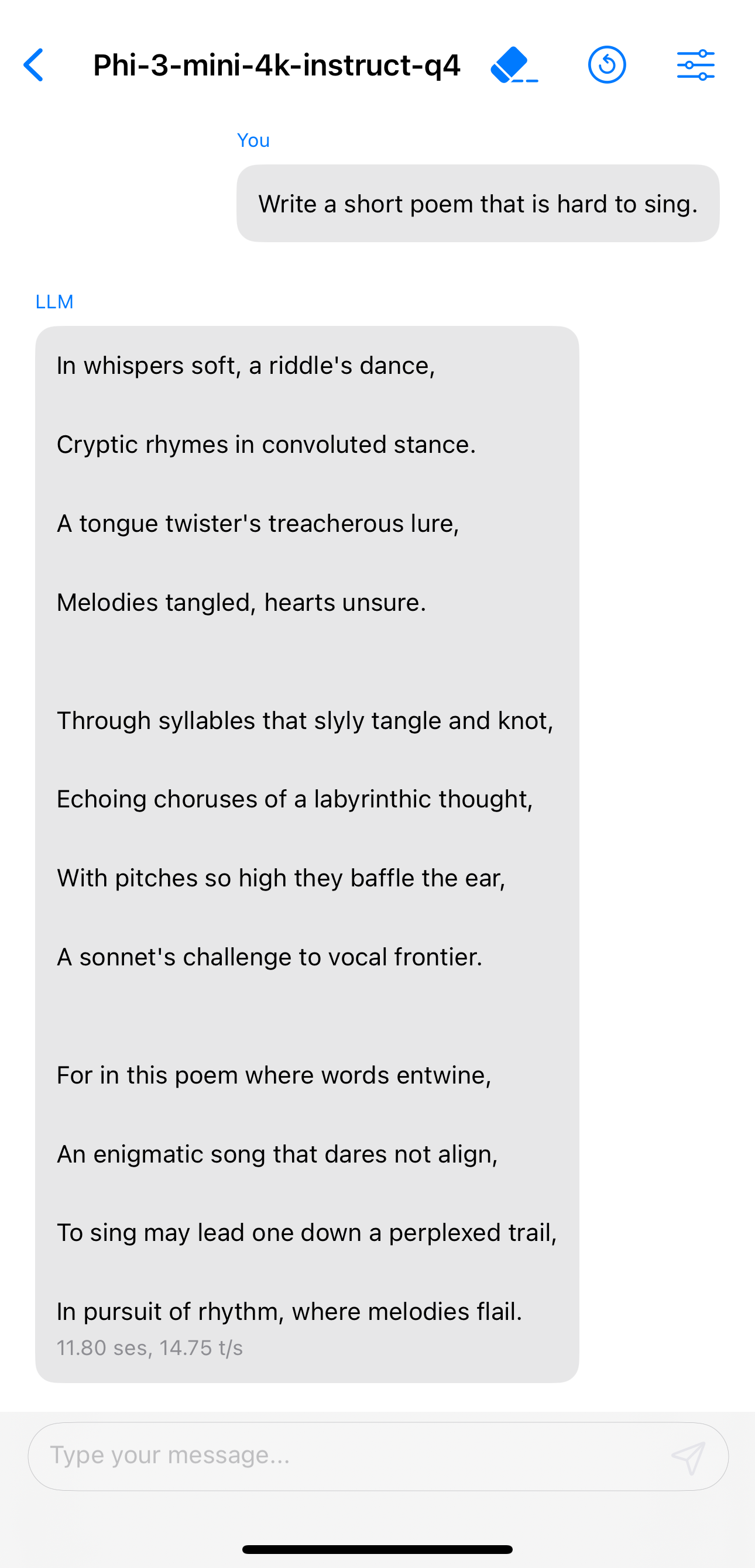}
\includegraphics[width=0.30\textwidth]{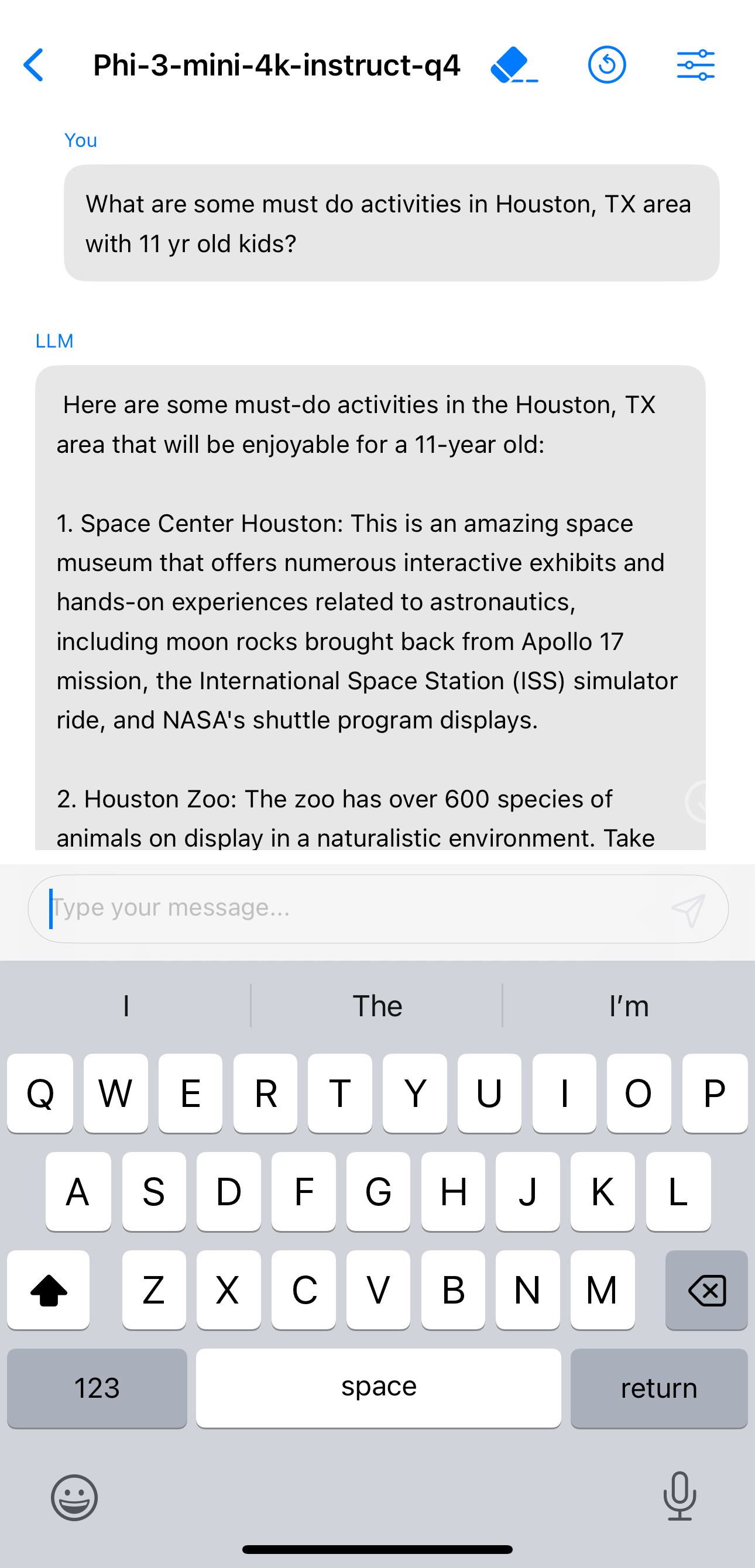}
\includegraphics[width=0.30\textwidth]{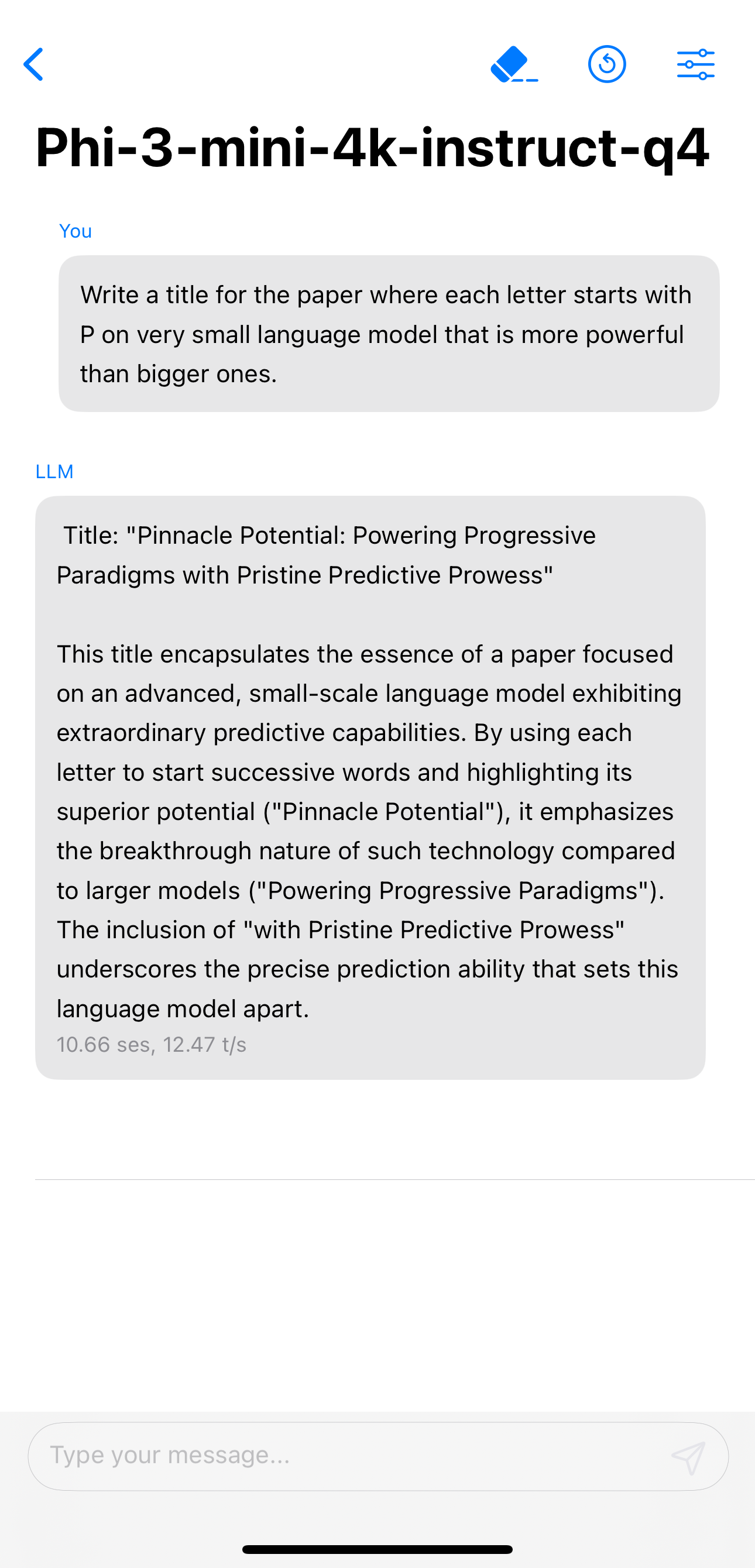}
    \caption{4-bit quantized \textbf{phi-3-mini} running natively on an iPhone with A16 Bionic chip, generating over 12 tokens per second.}
    \label{fig:1}
\end{figure}

The striking progress of AI in the last few years can be largely attributed to major efforts throughout the world towards {\em scaling-up} to ever-larger models and datasets. Large Language Models (LLMs) have steadily increased in size from a mere billion parameters just five years ago (GPT-2 had 1.5 billion parameters \cite{radford2019language}) to trillion parameters today. The impetus for this effort originates in the seemingly predictable improvement one obtains by training large models, the so-called {\em scaling laws} \cite{kaplan2020scaling, hoffmann2022training,muennighoff2023scaling}. However these laws assume a ``fixed'' data source. This assumption is now significantly disrupted by the existence of frontier LLMs themselves, which allow us to interact with data in novel ways. In our previous works on the phi models \cite{gunasekar2023textbooks,li2023textbooks, javaheripi2023phi} it was shown that a combination of LLM-based filtering of publicly available web data, and LLM-created synthetic data, enable performance in smaller language models that were typically seen only in much larger models. For example our previous model trained on this data recipe, \textbf{phi-2} (2.7B parameters), matched the performance of models $25$ times larger trained on regular data. In this report we present a new model, \textbf{phi-3-mini} (3.8B parameters), trained for 3.3T tokens on larger and more advanced versions of the datasets used in \textbf{phi-2}. With its small size, \textbf{phi-3-mini} can easily be inferenced locally on a modern phone (see Figure \ref{fig:1}), yet it achieves a quality that seems on-par with models such as Mixtral 8x7B \cite{jiang2024mixtral} and GPT-3.5.

\begin{AIbox}{}
\tt \footnotesize 
\begin{minipage}[t]{0.99\linewidth}
\textbf{User:} Explain why it is surprising that one can build a language model small enough to fit on a phone, yet almost as powerful as ChatGPT. Just use one funny sentence.
\end{minipage}
\\ \\
\begin{minipage}[t]{0.99\linewidth}
\textbf{phi-3-mini:} It's like fitting a supercomputer in a flip phone, but instead of breaking the phone, it just breaks the internet with its tiny, yet mighty, linguistic prowess!
\end{minipage}
\\ \\
\begin{minipage}[t]{0.99\linewidth}
\textbf{User:}  Okay now more serious answer, and note that this was achieved solely by changing the training data.
\end{minipage}
\\ \\
\begin{minipage}[t]{0.99\linewidth}
\textbf{phi-3-mini:} The development of a compact language model that rivals the capabilities of ChatGPT, while fitting on a phone, is a testament to the power of data-driven machine learning. By meticulously curating and optimizing the training dataset, researchers can significantly reduce the model's size without compromising its performance. [...]
\end{minipage}
\end{AIbox}

\section{Technical Specifications}
The \textbf{phi-3-mini} model is a transformer decoder architecture \cite{Vas17}, with default context length $4K$. We also introduce a long context version via LongRope \cite{ding2024longrope} that extends the context length to $128K$, called \textbf{phi-3-mini-128K}. 

To best benefit the open source community, \textbf{phi-3-mini} is built upon a similar block structure as Llama-2 \cite{touvron2023llama} and uses the same tokenizer with vocabulary size of 32064\footnote{We remove BoS tokens and add some additional tokens for chat template.}. {This means that all packages developed for Llama-2 family of models can be directly adapted to \textbf{phi-3-mini}}. The model uses $3072$ hidden dimension, $32$ heads and $32$ layers. We trained using bfloat16 for a total of 3.3T tokens. The model is already chat-finetuned, and the chat template is as follows:
\begin{AIbox}{}
\tt \footnotesize 
<|user|>$\backslash$n
Question
<|end|>$\backslash$n
<|assistant|>
\end{AIbox}

\begin{figure}[!h]
    \centering
    \includegraphics[scale=0.3]{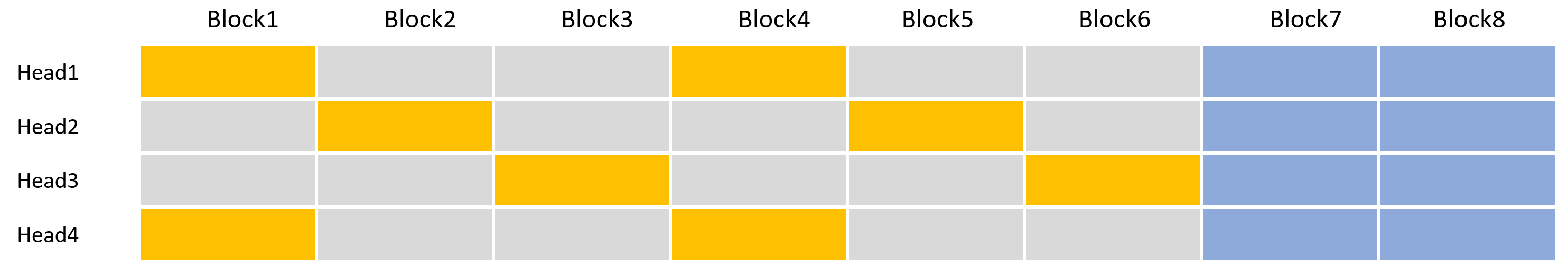}
    \caption{Toy illustration of the blocksparse attention in phi-3-small with 2 local blocks and vertical stride of 3. The table shows the Keys/values a query token in block 8 attended to. \textcolor{blue}{Blue}=local blocks, \textcolor{orange}{orange}=remote/vertical blocks, \textcolor{gray}{gray}=blocks skipped.}
    \label{fig:bs-atn-illustration}
\end{figure}

The \textbf{phi-3-small} model (7B parameters) leverages the tiktoken tokenizer (for better multilingual tokenization) with a vocabulary size of 100352\footnote{We remove unused tokens from the vocabulary.} and has default context length $8192$. 
It follows the standard decoder architecture of a 7B model class, having $32$ heads, $32$ layers and a hidden size of $4096$. We switched to GEGLU activation and used Maximal Update Parametrization (muP) \cite{yang2022tensor} to tune hyperparameters on a small proxy model and transfer them to the target 7B model. Those helped ensure better performance and training stability. 
Also, the model leverages a grouped-query attention, with $4$ queries sharing $1$ key. 
To optimize the training and inference speed, we design a novel blocksparse attention module.
For each attention head, the blocksparse attention enforces different sparsity patterns over KV cache. This ensures that all tokens are attended to on different heads for the given choice of sparsity.
As illustrated in Figure \ref{fig:bs-atn-illustration}, the context is then efficiently divided and conquered among attention heads, with significant KV cache reduction.
To achieve actual deployment speed-up from the blocksparse design, we implemented highly efficient, yet flexible kernels for both training and inference.
For training, we build a triton kernel based on Flash Attention \cite{dao2022flashattention}.
For inference, we implemented a kernel for the prefilling phase and extended the
paged attention kernel in vLLM for the decoding phase \cite{kwon2023efficient}.
%Both the training kernel and the inference kernel are implemented in a flexible way that allows users to freely conduct training and inference with arbitrary sparse patterns.
%todo, maybe move flexibility part after this
Lastly, in \textbf{phi-3-small} architecture, we alternate dense attention layers and blocksparse attention layers to optimize KV cache savings  while maintaining long context retrieval performance.
An additional 10\% multilingual data was also used for this model.

The \textbf{phi-3.5-MoE} adopts an Mixture-of-Experts (MoE) architecture to selectively activate parts of
modules on specific inputs to improve the model efficiency. It incorporates
MoE layer as its feedforward models, employing the top2 routing among 16 expert networks.
Particularly, each expert network is a separate GLU network and the routing module will
selectively activate 2 expert networks out of the 16 expert networks for each token, leaving
 16×3.8B model to have 6.6B activated parameters with 42B total parameters. Additionally, we utilize the SparseMixer approach \cite{Liu2023SparseMixer, Liu2023BridgingDA} for training the sparse router in the MoE model. For comparison with other Phi series models, \textbf{phi-3.5-MoE} uses the same tokenizer as \textbf{phi-3-medium} and \textbf{phi-3-mini} with vocabulary size of 32064.  
 
\paragraph{Highly capable language model running locally on a cell-phone.} Thanks to its small size, \textbf{phi-3-mini} can be quantized to 4-bits so that it only occupies $\approx$ 1.8GB of memory. We tested the quantized model by deploying \textbf{phi-3-mini} on iPhone 14 with A16 Bionic chip running natively on-device and fully offline achieving more than $12$ tokens per second.

\paragraph{Training Methodology.} We follow the sequence of works initiated in ``Textbooks Are All You Need''~\cite{gunasekar2023textbooks}, which utilize high quality training data to improve the performance of small language models and deviate from the standard {\em scaling-laws}. In this work we show that such method allows to reach the level of highly capable models such as GPT-3.5 or Mixtral with only 3.8B total parameters (while Mixtral has 45B total parameters for example). Our training data of consists of heavily filtered publicly available web data (according to the ``educational level'') from various open internet sources, as well as synthetic LLM-generated data. Pre-training is performed in two disjoint and sequential phases; phase-1 comprises mostly of web sources aimed at teaching the model general knowledge and language understanding. Phase-2 merges even more heavily filtered webdata (a subset used in Phase-1) with some synthetic data that teach the model logical reasoning and various niche skills. 

\begin{figure}
    \centering
    \includegraphics[width=0.9\textwidth]{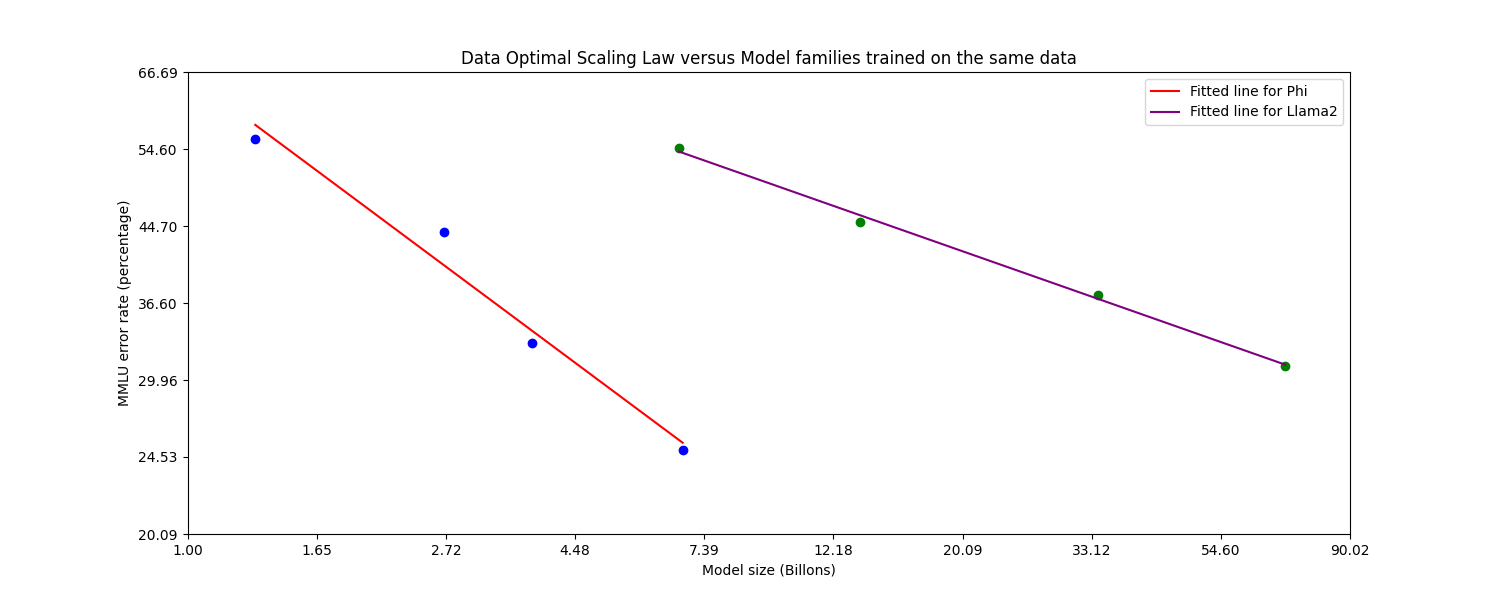}
    \caption{Scaling law close to the ``Data Optimal Regime" (from left to right: phi-1.5, phi-2, phi-3-mini, phi-3-small) versus Llama-2 family of models (7B, 13B, 34B, 70B) that were trained on the same fixed data. We plot the log of MMLU error versus the log of model size.}
    \label{fig:enter-label}
\end{figure}

\paragraph{Data Optimal Regime.} Unlike prior works that train language models in either ``compute optimal regime'' \cite{hoffmann2022training} or ``over-train regime'', we mainly focus on the quality of data for a {\em given scale}.\footnote{Just like for ``compute optimal regime", we use the term ``optimal" in an aspirational sense for ``data optimal regime". We are not implying that we actually found the provably ``optimal" data mixture for a given scale.} We try to calibrate the training data to be closer to the ``data optimal'' regime for small models. In particular, we filter the publicly available web data to contain the correct level of ``knowledge" and keep more web pages that could potentially improve the ``reasoning ability" for the model. As an example, the result of a game in premier league in a particular day might be good training data for frontier models, but we need to remove such information to leave more model capacity for ``reasoning'' for the mini size models. We compare our approach with Llama-2 in Figure~\ref{fig:enter-label}.

To test our data on larger size of models, we also trained \textbf{phi-3-medium}, a model with 14B parameters using the  same tokenizer and architecture of \textbf{phi-3-mini}, and trained on the same data for slightly more epochs (4.8T tokens total as for \textbf{phi-3-small}. The model has 40 heads and 40 layers, with embedding dimension 5120. We observe that some benchmarks improve much less from 7B to 14B than they do from 3.8B to 7B, perhaps indicating that our data mixture needs further work to be in the ``data optimal regime" for 14B parameters model.

\paragraph{Post-training.}
Post-training of \textbf{phi-3} went through  two stages, including supervised finetuning (SFT) and direct preference  optimization (DPO). SFT leverages highly curated  high-quality data across diverse domains, e.g., math, coding, reasoning, conversation, model identity, and safety. The SFT data mix starts with using English-only examples. DPO data covers chat format data, reasoning, and responsible AI (RAI) efforts. We use DPO to steer the model away from unwanted behavior, by using those outputs as “rejected” responses. Besides improvement in math, coding, reasoning, robustness, and safety, post-training transforms a language model to an AI assistant that users can efficiently and safely interact with.

\section{Academic benchmarks}

On the next page we report the results for \textbf{phi-3} on standard open-source benchmarks measuring the model's reasoning ability (both common sense reasoning and logical reasoning). We compare to phi-2 \cite{javaheripi2023phi}, Mistral-7b-v0.1 \cite{jiang2023mistral}, Mixtral-8x7b \cite{jiang2024mixtral}, Gemma 7B \cite{gemmateam2024gemma}, Llama-3-instruct-8b \cite{llama3}, and GPT-3.5. All the reported numbers are produced with the exact same pipeline to ensure that the numbers are comparable. These numbers might differ from other published numbers due to slightly different choices in the evaluation. As is now standard, we use few-shot prompts to evaluate the models, at temperature $0$. The prompts and number of shots are part of a Microsoft internal tool to evaluate language models, and in particular we did no optimization to the pipeline for the \textbf{phi-3} models.\footnote{For example, we found that using \#\# before the Question can lead to a noticeable improvement to \textbf{phi-3-mini}'s results across many benchmarks, but we did not do such changes in the prompts.} 
The number of $k$--shot examples is listed per-benchmark. 
An example of a 2-shot prompt is described in Appendix \ref{sec:prompt}.

\begin{center}
\begin{adjustbox}{width=0.95\textwidth,center}
\begin{tabular}{c||ccccccccc } 
\label{tbl:benchmarks}
&\makecell{Phi-3-mini\\ \footnotesize 3.8b } & \makecell{Phi-3-small\\ \footnotesize 7b } &  \makecell{Phi-3-medium\\ \footnotesize 14b } & %\makecell{Phi-3-MoE\\ \footnotesize 16x3.8b} &
\makecell{Phi-2 \\ \footnotesize 2.7b } & \makecell{Mistral\\ \footnotesize 7b } &\makecell{Gemma \\ \footnotesize 7b }&\makecell{Llama-3-In \\ \footnotesize 8b }  & \makecell{Mixtral\\ \footnotesize 8x7b }   &  \makecell{GPT-3.5 \\ \footnotesize version 1106}  \\
\hline & \\[-1.5ex]

\datasetcell{MMLU}{5-Shot}{\cite{hendrycks2021measuring} }         & 68.8 & 75.7 & 78.0 &% 79.4 & 
56.3 & 61.7& 63.6  & 66.5 & 70.5  & 71.4  \\

\datasetcell{HellaSwag}{5-Shot}{\cite{zellers2019hellaswag} }      & 76.7& 77.0 & 82.4& %83.7 & 
53.6 & 58.5 & 49.8 & 71.1  & 70.4  & 78.8 \\ 
\datasetcell{ANLI}{7-Shot}{\cite{nie2020adversarial}}                                       & 52.8 & 58.1 &55.8 % & 60.6
& 42.5 & 47.1 &  48.7 & 57.3  & 55.2 & 58.1  \\
\hline & \\[-1.5ex]
\datasetcell{ GSM-8K}{8-Shot; CoT}{\cite{cobbe2021training} }      & 82.5 & 89.6  & 91.0&% 90.4 &
61.1 & 46.4 &  59.8  & 77.4 & 64.7 & 78.1  \\ 

\datasetcell{ MATH}{0-Shot; CoT}{\cite{hendrycksmath2021} }      & 41.3 & 34.6  & 53.1 &% 58.9 &
-- & 15.0 &  13.6  & 28.2 & 11.1 & 45.3  \\ 

\hline
\datasetcell{ MedQA}{2-Shot}{\cite{jin2020disease} }                                    &53.8& 65.4 & 69.9 % & 70.4 
& 40.9 & 50.0  & 49.6  & 60.5 & 62.2&  63.4  \\ 
\datasetcell{ AGIEval}{0-Shot}{\cite{zhong2023agieval} }           & 37.5 &45.1  & 50.2 &% 48.2 & 
29.8 & 35.1  & 42.1  & 42.0 & 45.2  & 48.4  \\ 
\datasetcell{ TriviaQA}{5-Shot}{ \cite{joshi2017triviaqa}}                                 & 64.0 & 58.1 &73.9% & 73.9
& 45.2 & 75.2  & 72.3 & 67.7   &  82.2 &  85.8 \\ 
\hline & \\[-1.5ex]
\datasetcell{Arc-C}{10-Shot}{\cite{clark2018think} }               & 84.9 & 90.7 & 91.6% & 92.0
& 75.9 & 78.6 & 78.3  & 82.8 & 87.3& 87.4 \\ 
\datasetcell{Arc-E}{10-Shot}{\cite{clark2018think} }               & 94.6 & 97.0& 97.7&% 98.0 &
88.5 & 90.6 & 91.4  & 93.4 & 95.6 & 96.3  \\ 
\datasetcell{ PIQA}{5-Shot}{\cite{bisk2019piqa} }                  & 84.2 &86.9 &87.9 &% 89.0 &
60.2 & 77.7 & 78.1  & 75.7 & 86.0& 86.6  \\ 
\datasetcell{ SociQA}{5-Shot}{\cite{bisk2019piqa} }                & 76.6 & 79.2 & 80.2% & 79.5
&68.3 &  74.6 & 65.5 & 73.9  & 75.9 & 68.3  \\ 
\hline & \\[-1.5ex]

\datasetcell{ BigBench-Hard}{3-Shot; CoT}{\cite{srivastava2022beyond,suzgun2022challenging} }    
                                                                   & 71.7 & 79.1 & 81.4 
                                                  %&   81.4  
                                                  & 59.4 & 57.3  & 59.6  & 51.5 & 69.7 & 68.32 \\ 
\datasetcell{WinoGrande}{5-Shot}{\cite{sakaguchi2019winogrande} }  & 70.8 & 81.5 & 81.5% & 81.4
& 54.7 & 54.2 & 55.6 & 65.0 & 62.0  & 68.8  \\ 
\datasetcell{OpenBookQA}{10-Shot}{\cite{mihaylov2018suit} }        & 83.2 & 88.0 & 87.4 &% 89.8 &
73.6 & 79.8 & 78.6  & 82.6 & 85.8  & 86.0  \\ 
\datasetcell{BoolQ}{2-Shot}{\cite{clark2019boolq} }                & 77.2 & 84.8  & 86.5 &% 83.4 &
--& 72.2 & 66.0 & 80.9 &77.6& 79.1  \\ % misantac boolQ incorrectly 77.2 for phi-3-mini on 1st upload, should be 77.6
\datasetcell{CommonSenseQA}{10-Shot}{\cite{talmor2019commonsenseqa} }  & 80.2& 80.0 &82.8 &% 81.8 &
69.3 &  72.6 & 76.2 & 79.0 & 78.1 & 79.6  \\ 
\datasetcell{TruthfulQA}{10-Shot; MC2}{\cite{lin2022truthfulqa} }       & 65.0 & 70.2 & 75.1 &% 74.5 &
--& 53.0 & 52.1  & 63.2 & 60.1  & 85.8  \\ 
%BoolQ \cite{clark2019boolq} & --- & --- & --- & --- & ---  \\ 

\hline & \\[-1.5ex]
\datasetcell{ HumanEval}{0-Shot}{\cite{chen2021evaluating} }       & 58.5& 61.0 & 62.2% & 74.4
& 59.0 & 28.0  & 34.1  & 60.4 & 37.8 & 62.2 \\ 
\datasetcell{ MBPP}{3-Shot}{\cite{austin2021program} }             & 70.0 & 71.7 & 75.2% & 80.3
& 60.6 & 50.8 & 51.5  & 67.7 & 60.2 & 77.8  \\ 
\hline & \\[-1.5ex]
Average                                                            & 69.7 & 73.6 & 76.7 &% 78.5 &
-- & 58.9 & 59.3  & 67.3 & 66.8 & 72.8  \\   % phi-small is 
\hline & \\[-1.5ex]
\datasetcell{GPQA}{2-Shot; CoT}{\cite{rein2023gpqa}}                                       & 32.8 & 34.3  & --&% 37.9 &
--& --&  --  &-- & -- & 29.0 \\ 
\datasetcell{MT Bench}{2 round ave.}{\cite{zheng2023judging}}  & 8.38 & 8.70 & 8.91% & 8.86
& --& --&  --  &--  & -- & 8.35  \\

\iffalse
llama-3-70b	
78.2	mmlu
80.0	hella
61.8	anLi
83.7	gsm8k
75.3	medqa
57.3	agieval
85.1	trivia
92.4	arc c
98.0	arc e 
89.3	piqa
78.2	siqa
79.7	bbh
77.7	winogr
92.9	openbookqa
82.7	boolq
84.4	commonsenseqa
55.4	truthqa
40.2	humane
74.9	mbpp
	
77.22105263	average
\fi

\end{tabular}
\end{adjustbox}
\end{center}

\section{Multilingual and Long Context}

To enhance the Phi-3 models with multilingual and long-context capabilities, we developed the versions \textbf{phi-3.5-mini} and \textbf{phi-3.5-MoE}, which incorporate more multilingual and long-text data during mid-training. Specifically, we employed the long-rope method \cite{ding2024longrope} and a mixed context window approach to expand the context length limit from 4K to 128K without compromising performance on 4K-context tasks.

\begin{figure}[h]
    \centering
    \includegraphics[width=0.9\textwidth]{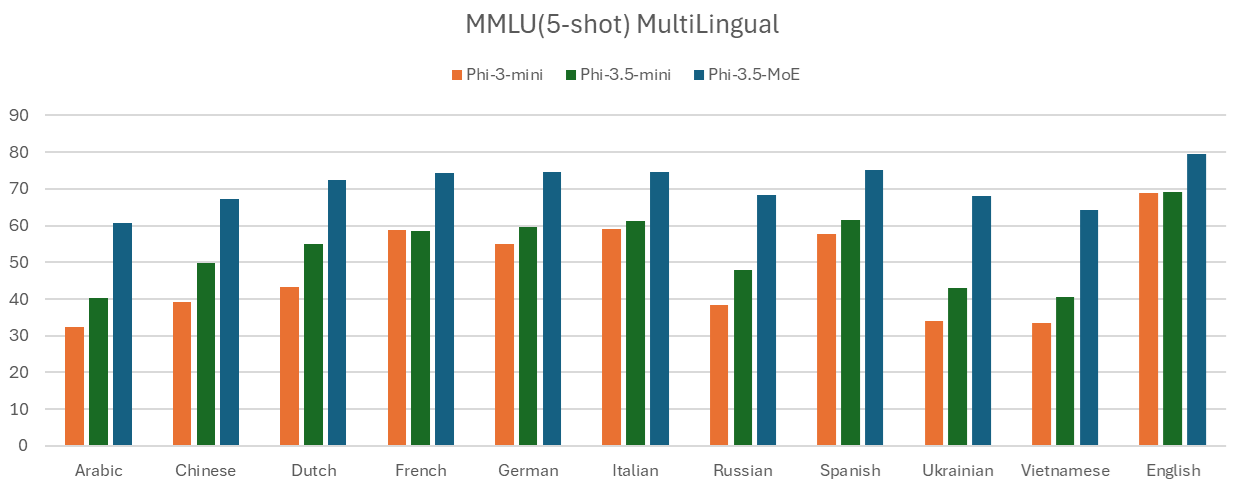}
    \caption{Comparison of \textbf{phi-3-mini}, \textbf{phi-3.5-mini} and \textbf{phi-3.5-MoE} on MMLU-Multilingual tasks }
    \label{fig:ml_moe}
\end{figure}

Figure ~\ref{fig:ml_moe} compares the performance of \textbf{phi-3-mini}, \textbf{phi-3.5-mini}, and \textbf{phi-3.5-MoE} on MMLU multilingual tasks. \textbf{phi-3.5-mini} demonstrates significant improvement over \textbf{phi-3-mini} in languages such as Arabic, Chinese, Russian, Ukrainian, and Vietnamese, with average MMLU-multilingual scores of $55.4$ and $47.3$, respectively. Due to its larger model capacity, \textbf{phi-3.5-MoE} achieves a significantly higher average score of $69.9$, outperforming \textbf{phi-3.5-mini}.

We evaluate the \textbf{phi-3.5-mini} and \textbf{phi-3.5-MoE} models on two long-context understanding tasks: RULER \cite{hsieh2024rulerwhatsrealcontext} and RepoQA \cite{liu2024repoqaevaluatinglongcontext}. As shown in Tables \ref{tbl:longrepoqa} and \ref{tbl:longruler}, both \textbf{phi-3.5-MoE} and \textbf{phi-3.5-mini} outperform other open-source models with larger sizes, such as Llama-3.1-8B, Mixtral-8x7B, and Mixtral-8x22B, on the RepoQA task, and achieve comparable performance to Llama-3.1-8B on the RULER task. However, we observe a significant performance drop when testing the 128K context window on the RULER task. We suspect this is due to the lack of high-quality long-context data in mid-training, an issue we plan to address in the next version of the model release.  

 In the table \ref{tab:benchmark-comparison-3.5}, we present a detailed evaluation of the \textbf{phi-3.5-mini} and \textbf{phi-3.5-MoE} models compared with recent SoTA pretrained language models, such as GPT-4o-mini, Gemini-1.5 Flash, and open-source models like Llama-3.1-8B and the Mistral models. The results show that \textbf{phi-3.5-mini} achieves performance comparable to much larger models like Mistral-Nemo-12B and Llama-3.1-8B, while \textbf{phi-3.5-MoE} significantly outperforms other open-source models, offers performance comparable to Gemini-1.5 Flash, and achieves above 90\% of the average performance of GPT-4o-mini across various language benchmarks.

\begin{table}[t]
\begin{center}
\begin{adjustbox}{width=0.7\textwidth,center}
\begin{tabular}{ cc||cccccc } 
Model & Ctx Size & Python & C++ & Rust & Java & TypeScript & Average  \\
\hline & \\[-1.5ex]
gpt-4O-2024-05-13 & 128k & 95 & 80 & 85 & 96 & 97 & 90.6 \\
gemini-1.5-flash-latest & 1000k & 93 & 79 & 87 & 94 & 97 & 90 \\
\textbf{Phi-3.5-MoE} & 128k & 89 & 74 & 81 & 88 & 95 & 85 \\
\textbf{Phi-3.5-Mini} & 128k & 86 & 67 & 73 & 77 & 82 & 77 \\
Llama-3.1-8B-Instruct & 128k & 80 & 65 & 73 & 76 & 63 & 71 \\
Mixtral-8x7B-Instruct-v0.1 & 32k & 66 & 65 & 64 & 71 & 74 & 68 \\
Mixtral-8x22B-Instruct-v0.1 & 64k & 60 & 67 & 74 & 83 & 55 & 67.8 \\
\end{tabular}
\end{adjustbox}
\end{center}
\caption{Comparison results on RepoQA benchmark.}
\label{tbl:longrepoqa}
\end{table}

\begin{table}[t]
\begin{center}
\begin{adjustbox}{width=0.7\textwidth,center}
\begin{tabular}{ cc||ccccccc } 
Model & Ctx Size & 4k & 8k & 16k & 32k & 64k & 128k & Average  \\
\hline & \\[-1.5ex]
Llama-3.1-8B-Instruct & 128k & 95.5 & 93.8 & 91.6 & 87.4 & 84.7 & 77.0 & 88.3 \\
\textbf{Phi-3.5-MoE} & 128k & 94.8 & 93.0 & 93.2 & 91.6 & 85.7 & 64.2 & 87.1 \\
\textbf{Phi-3.5-Mini} & 128k & 94.3 & 91.1 & 90.7 & 87.1 & 78.0 & 63.6 & 84.1 \\
Mixtral-8x22B-Instruct-v0.1 & 64k & 95.6 & 94.9 & 93.4 & 90.9 & 84.7 & 31.7 & 81.9 \\
Mixtral-8x7B-Instruct-v0.1 & 32k &94.9 & 92.1 &	92.5 &85.9 &72.4 & 44.5 & 80.4 \\
\end{tabular}
\end{adjustbox}
\end{center}
\caption{Comparison results on RULER benchmark.}
\label{tbl:longruler}
\end{table}

\begin{table}[t]
\begin{center}
\begin{adjustbox}{width=1.0\textwidth,center}
\begin{tabular}{ c|c||cccccccc } 
\textbf{Category} & \textbf{Benchmark} & \makecell{Phi-3.5-mini \\ \footnotesize 3.8B}  & \makecell{Phi-3.5-MoE \\ \footnotesize 16x3.8B}  & 
\makecell{Mistral \\ \footnotesize 7B} & \makecell{Mistral-Nemo \\ \footnotesize 12B} & \makecell{Llama-3.1-In\\ \footnotesize 8B} & \makecell{Gemma-2 \\ \footnotesize 9B} & \makecell{Gemini-1.5 \\ \footnotesize Flash} & \makecell{GPT-4o-mini} \\ \hline
\multirow{2}{*}{Popular} & Arena Hard & 37 & 37.9 & 18.1 & 39.4 & 25.7 & 42 & 55.2 & 75 \\ %\cline{2-9} 
 & \makecell{BigBench Hard \\ \footnotesize CoT (0-shot)} & 69 & 79.1 & 33.4 & 60.2 & 63.4 & 63.5 & 66.7 & 80.4 \\ \hline
\multirow{2}{*}{MMLU} & \makecell{MMLU \\ \footnotesize (5-shot)} & 69 & 78.9 &  60.3 & 67.2 & 68.1 & 71.3 & 78.7 & 77.2 \\ %\cline{2-9} 
 & \makecell{MMLU-Pro \\ \footnotesize (0-shot, CoT)} & 47.5 & 54.3 & 18 & 40.7 & 44 & 50.1 & 57.2 & 62.8 \\ \hline
\multirow{9}{*}{Reasoning} & \makecell{ARC Challenge \\ \footnotesize (10-shot)} & 84.6 & 91.0 & 77.9 & 84.8 & 83.1 & 89.8 & 92.8 & 93.5 \\ %\cline{2-9} 
 & \makecell{ BoolQ \\ \footnotesize (2-shot) }& 78 & 84.6 & 80.5 & 82.5 & 82.8 & 85.7 & 85.8 & 88.7 \\ %\cline{2-9} 
 & \makecell{GPQA \\ \footnotesize (0-shot, CoT)}  & 27.2 & 36.8 & 15.6 & 28.6 & 26.3 & 29.2 & 37.5 & 41.1 \\ %\cline{2-9} 
 & \makecell{ HellaSwag \\ \footnotesize (5-shot) }& 69.4 & 83.8 & 71.6 & 76.7 & 73.5 & 80.9 & 67.5 & 87.1 \\ %\cline{2-9} 
 & \makecell{ OpenBookQA \\ \footnotesize (10-shot) } & 79.2 & 89.6 & 78 & 84.4 & 84.8 & 89.6 & 89 & 90 \\ %\cline{2-9} 
 & \makecell{ PIQA \\ \footnotesize (5-shot) } & 81 & 88.6 & 73.4 & 83.5 & 81.2 & 83.7 & 87.5 & 88.7 \\ %\cline{2-9} 
 & \makecell{ Social IQA \\ \footnotesize (5-shot) } & 74.7 & 78.0 & 73 & 75.3 & 71.8 & 74.7 & 77.8 & 82.9 \\ %\cline{2-9} 
 & \makecell{ TruthfulQA \\ \footnotesize (10-shot,MC2) } & 64 & 77.5 & 64.7 & 68.1 & 69.2 & 76.6 & 76.6 & 78.2 \\ %\cline{2-9} 
 & \makecell{ WinoGrande \\ \footnotesize (5-shot) } & 68.5 & 81.3 & 58.1 & 70.4 & 64.7 & 74 & 74.7 & 76.9 \\ \hline
\multirow{2}{*}{Multilingual} & \makecell{ Ml MMLU \\ \footnotesize (5-shot) } & 55.4 & 69.9 & 47.4 & 58.9 & 56.2 & 63.8 & 77.2 & 72.9 \\ %\cline{2-9} 
 & \makecell{ MGSM \\ \footnotesize (0-shot CoT) } & 47.9 & 58.7 & 31.8 & 63.3 & 56.7 & 76.4 & 75.8 & 81.7 \\ \hline
\multirow{2}{*}{Math} & \makecell{ GSM8K \\ \footnotesize (8-shot, CoT) } & 86.2 & 88.7 & 54.4 & 84.2 & 82.4 & 84.9 & 82.4 & 91.3 \\ %\cline{2-9} 
 & \makecell{ MATH \\ \footnotesize (0-shot, CoT) } & 48.5 & 59.5 & 19 & 31.2 & 47.6 & 50.9 & 38 & 70.2 \\ \hline
\multirow{2}{*}{Long context} & Qasper & 41.9 & 40.0 & 31.4 & 30.7 & 37.2 & 13.9 & 43.5 & 39.8 \\ %\cline{2-9} 
 & SQuALITY & 24.3 & 24.1 & 25.9 & 25.8 & 26.2 & 0 & 23.5 & 23.8 \\ \hline
\multirow{2}{*}{Code} & \makecell{ HumanEval \\ \footnotesize (0-shot)} & 61.5 & 70.7 & 35.4 & 63.4 & 66.5 & 61 & 74.4 & 86.6 \\ %\cline{2-9} 
 & \makecell{ MBPP \\ \footnotesize (3-shot) }& 68.6 & 80.8 & 50.4 & 68.1 & 69.4 & 69.3 & 77.5 & 84.1 \\ \hline
\multicolumn{2}{c}{Average} & 61.1 & {69.2} & {48.5} & {61.3} & {61.0} & {63.3} & {68.5} & {74.9} \\ %\hline
\end{tabular}
\end{adjustbox}
\caption{Model quality on representative benchmarks}
\label{tab:benchmark-comparison-3.5}
\end{center}
\end{table}

\section{Safety}
\begin{figure}[h]
    \centering
    \includegraphics[width=0.9\textwidth]{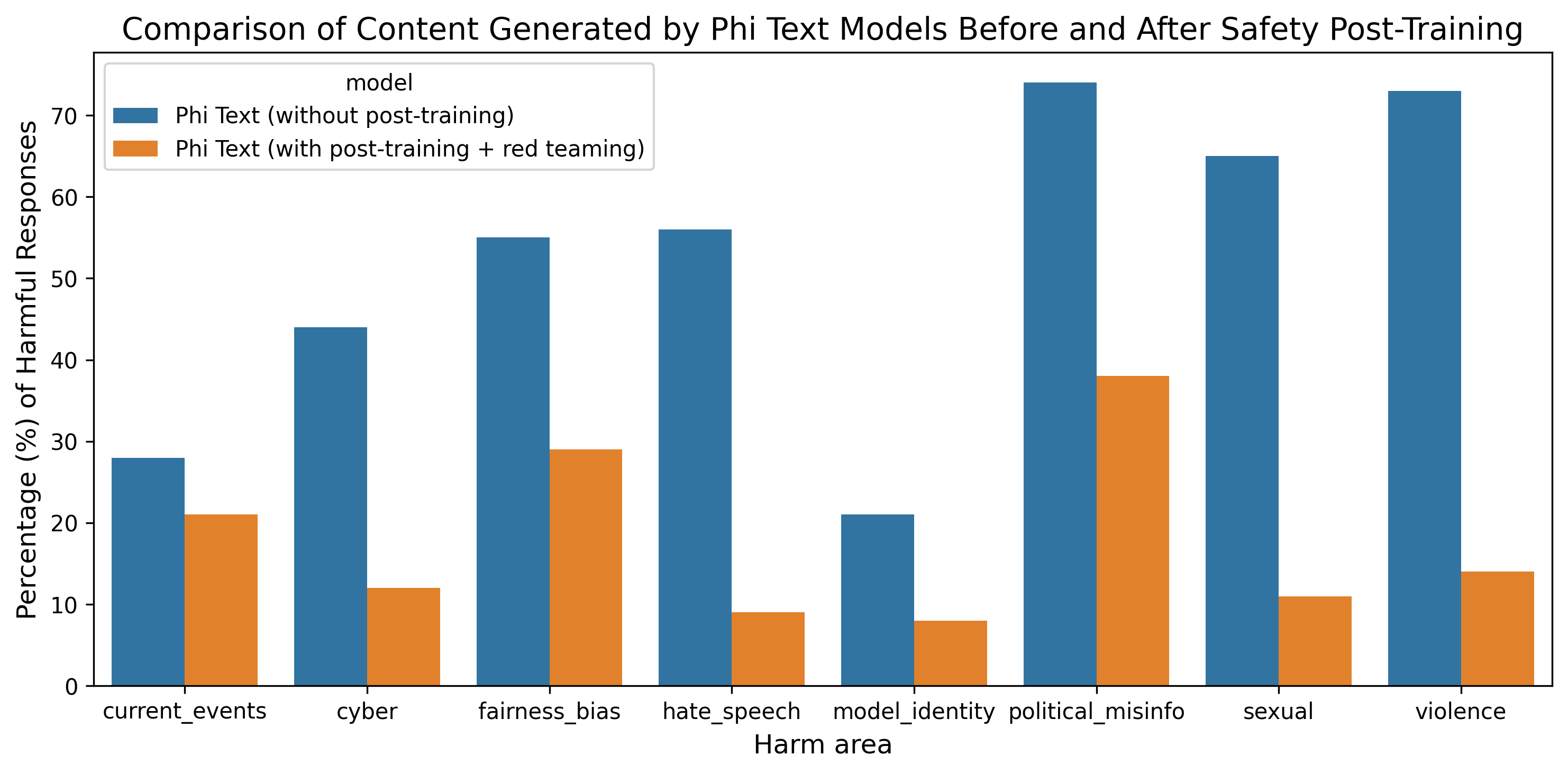}
    \caption{Comparison of harmful response percentages by Microsoft AI Red Team between \textbf{phi-3-mini} before and after the safety alignment. Note that the harmful response percentages in this chart are inflated numbers as the red team tried to induce \textbf{phi-3-mini} in an adversarial way to generate harmful responses through multi-turn conversations.}
    \label{fig:safety-pt}
\end{figure}

\textbf{Phi-3-mini} was developed in accordance with Microsoft’s responsible AI principles. The overall approach consisted of safety alignment in post-training, red-teaming, automated testing and evaluations across dozens of RAI harm categories. Helpfulness and harmlessness preference datasets \cite{bai2022training, ji2023beavertails} with modifications inspired by \cite{bianchi2024safetytuned} and multiple in-house generated datasets were leveraged to address the RAI harm categories in safety post-training. An independent red team at Microsoft iteratively examined \textbf{phi-3-mini} to further identify areas of improvement during the post-training process. Based on their feedback, we curated additional datasets tailored to address their insights, thereby refining the post-training dataset. This process resulted in significant decrease of harmful response rates, as shown in Figure \ref{fig:safety-pt}.

The safety alignment of \textbf{phi-3-small}, \textbf{phi-3-medium} and \textbf{phi-3.5-MoE} was conducted by undergoing the same red-teaming process, utilizing identical datasets, and incorporating a slightly larger number of samples. Table \ref{tab:rai-benchmarks} shows the results of in-house RAI benchmarks \cite{magooda2023framework} for \textbf{phi-3} models compared to phi-2 \cite{javaheripi2023phi}, Mistral-7b-v0.1 \cite{jiang2023mistral}, Gemma 7b \cite{gemmateam2024gemma}, and Llama-3-instruct-8b \cite{llama3}. This benchmark utilized GPT-4 to simulate multi-turn conversations in five different categories and to evaluate the model responses. Ungroundedness between 0 (fully grounded) and 4 (not grounded) measures if the information in a response is based on a given prompt. In other categories, responses were evaluated in terms of the severity of harmfulness from 0 (no harm) to 7 (extreme harm) and the defect rates (DR-$x$) were computed as the percentage of samples with the severity score being greater than or equal to $x$.

\begin{table}
\begin{center}
    \begin{adjustbox}{width=0.95\textwidth,center}
    \setlength\extrarowheight{6pt}
        \begin{tabular}{ c||cccccccc } 
        & \makecell{Phi-3-mini \\ \footnotesize 3.8b} & \makecell{Phi-3-small \\ \footnotesize 7b} & \makecell{Phi-3-medium \\ \footnotesize 14b} & \makecell{Phi-3.5-MoE \\ \footnotesize 16x3.8b} & \makecell{Phi-2 \\ \footnotesize 2.7b } & \makecell{Mistral\\ \footnotesize 7b } & \makecell{Gemma \\ \footnotesize 7b} & \makecell{Llama-3-In \\ \footnotesize 8b} \\
        \hline & \\[-3.5ex]
        Ungroundedness  & 0.603 & 0.299 & 0.213 & 0.228 & 1.481 & 0.935 & 0.679 & 0.328  \\
        Third Party Harm (DR-1) & 0.240 & 0.253 & 0.251 & 0.105 & 0.240 & 0.562 & 0.383 & 0.373 \\
        Harmful Content Continuation (DR-3) & 0.007 & 0.003 & 0.010 & 0.005 & 0.029 & 0.026 & 0.013 & 0.013 \\
        Harmful Content Summarization (DR-3) & 0.100 & 0.110 & 0.112 & 0.12 & 0.144 & 0.223 & 0.103 & 0.082 \\
        Jailbreak (DR-1) & 0.123 & 0.107 & 0.111 & 0.106 & 0.150 & 0.156 & 0.114 & 0.130 \\
        \end{tabular}
    \end{adjustbox}
\end{center}
\caption{Comparison of Microsoft internal multi-turn conversation RAI benchmark results of \textbf{phi-3} models and other models. Note that a lower value indicates a better performance for all metrics in the table.}
\label{tab:rai-benchmarks}
\end{table}

\section{Weakness}
\begin{figure}
    \centering
    \includegraphics[width=0.48\textwidth]{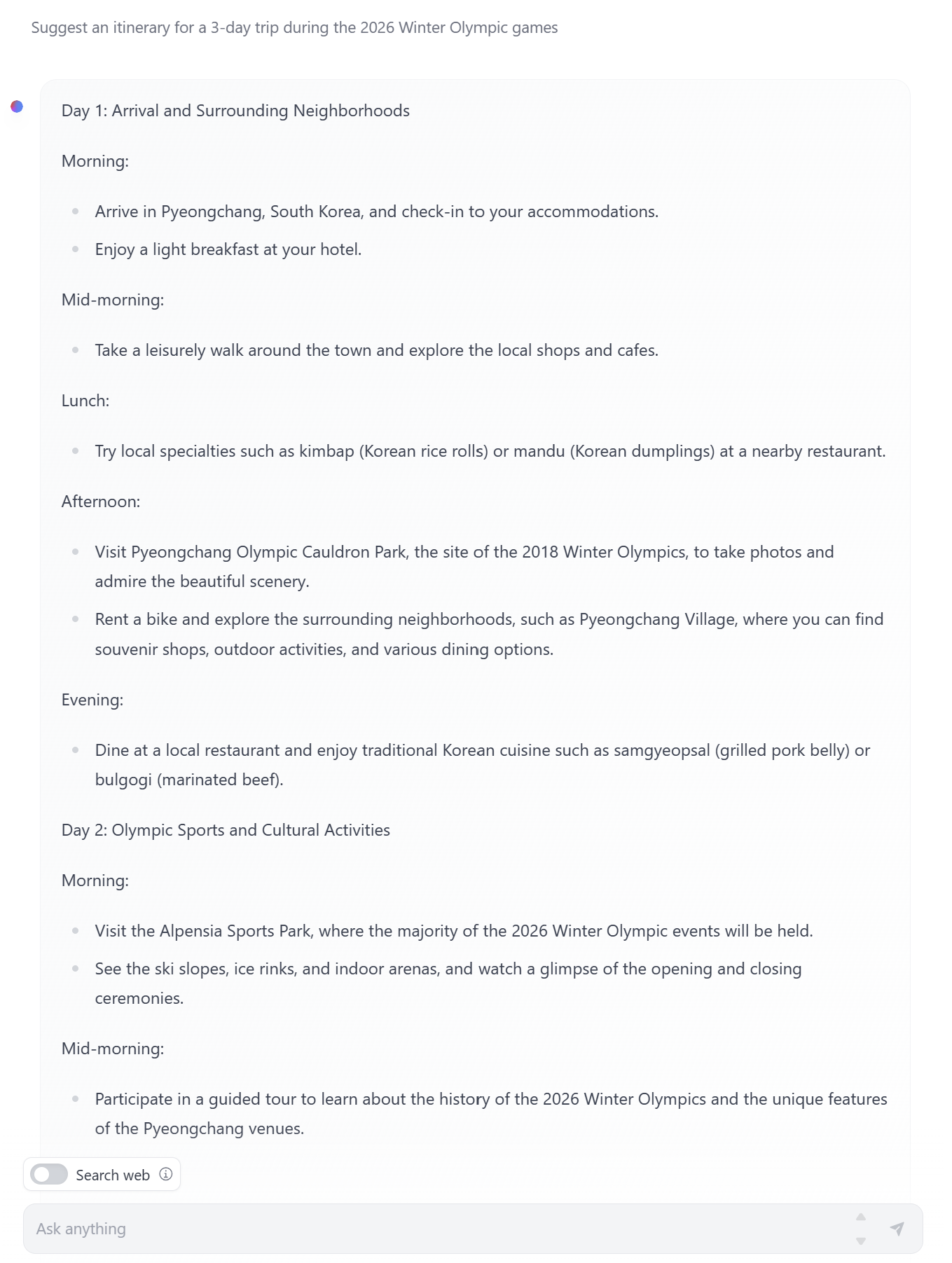}    \includegraphics[width=0.48\textwidth]{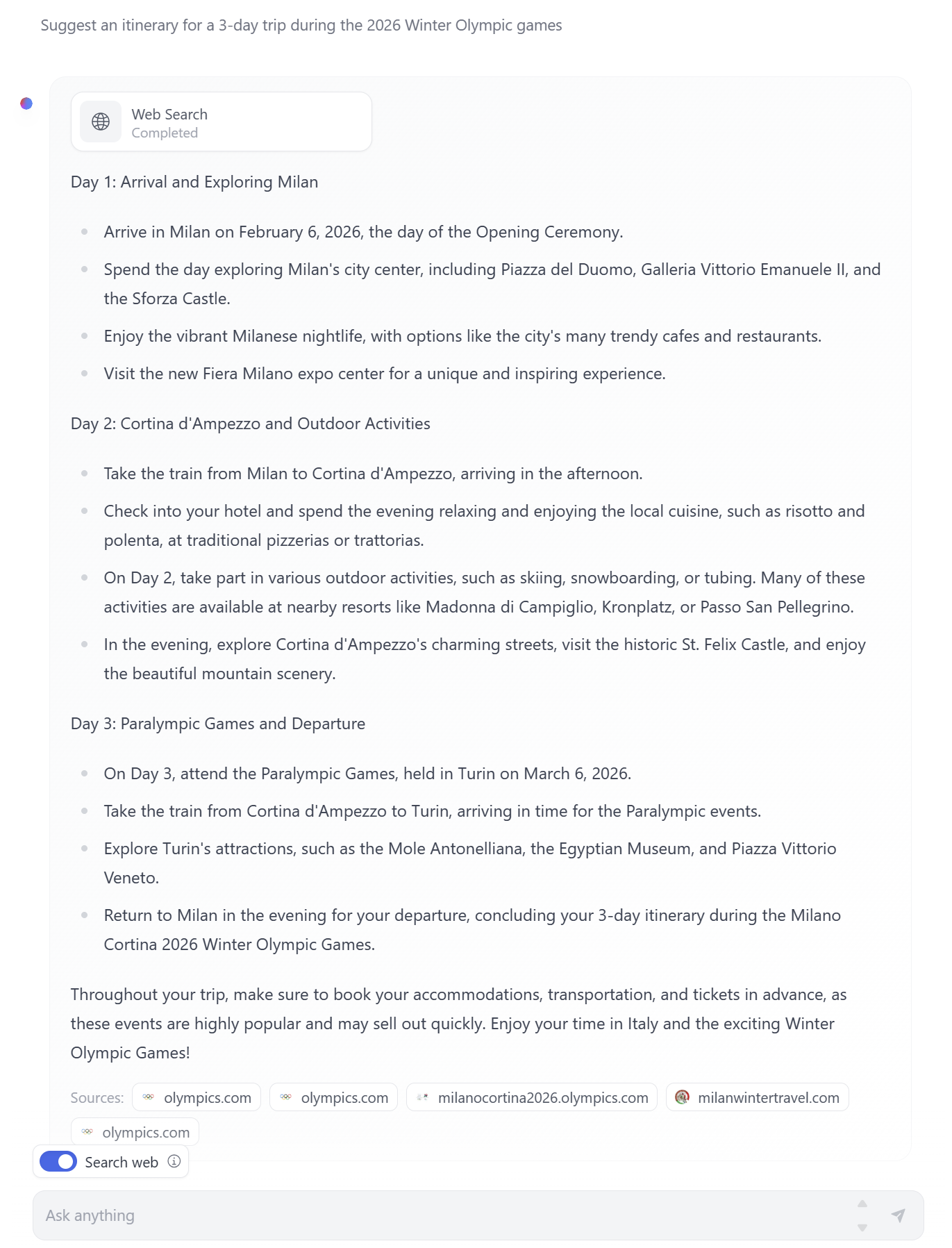}
    \caption{Left: \textbf{phi-3-mini}'s completion without search. Right: \textbf{phi-3-mini}'s completion with search, using the default HuggingFace Chat-UI search ability. For reference, the 2026 Winter Olympic Games are scheduled to be held in Milano and Cortina in Italy, while the 2022 and 2018 Winter Olympic Games were held in Beijing, China and PyeongChang, Korea, respectively. Without the search results, the response is incorrect, while with the web search, not only does the response become accurate, but also gets more specific with suggestions.}
    \label{fig:search}
\end{figure}

In terms of LLM capabilities, while $\textbf{phi-3-mini}$ model achieves similar level of language understanding and reasoning ability as much larger models, it is still fundamentally limited by its size for certain tasks. The model simply does not have the capacity to store too much ``factual knowledge'', which can be seen for example with low performance on TriviaQA.
However, we believe such weakness can be resolved by augmentation with a search engine. We show an example using the HuggingFace default Chat-UI with \textbf{phi-3-mini} in Figure \ref{fig:search}. Another weakness related to model's capacity is that we mostly restricted the language to English. Exploring multilingual capabilities for Small Language Models is an important next step, with some initial promising results on \textbf{phi-3-small} by including more multilingual data.

Despite our diligent RAI efforts, as with most LLMs, there remains challenges around factual inaccuracies (or hallucinations), reproduction or amplification of biases, inappropriate content generation, and safety issues. The use of carefully curated training data, and targeted post-training, and improvements from red-teaming insights significantly mitigates these issues across all dimensions. However, there is significant work ahead to fully  address these challenges, and downstream use of the models should be evaluated for the specific use cases and safety considerations for that context.

\section{Phi-3.5-Vision}
\begin{figure}[h]
    \centering
    \includegraphics[width=0.98\textwidth]{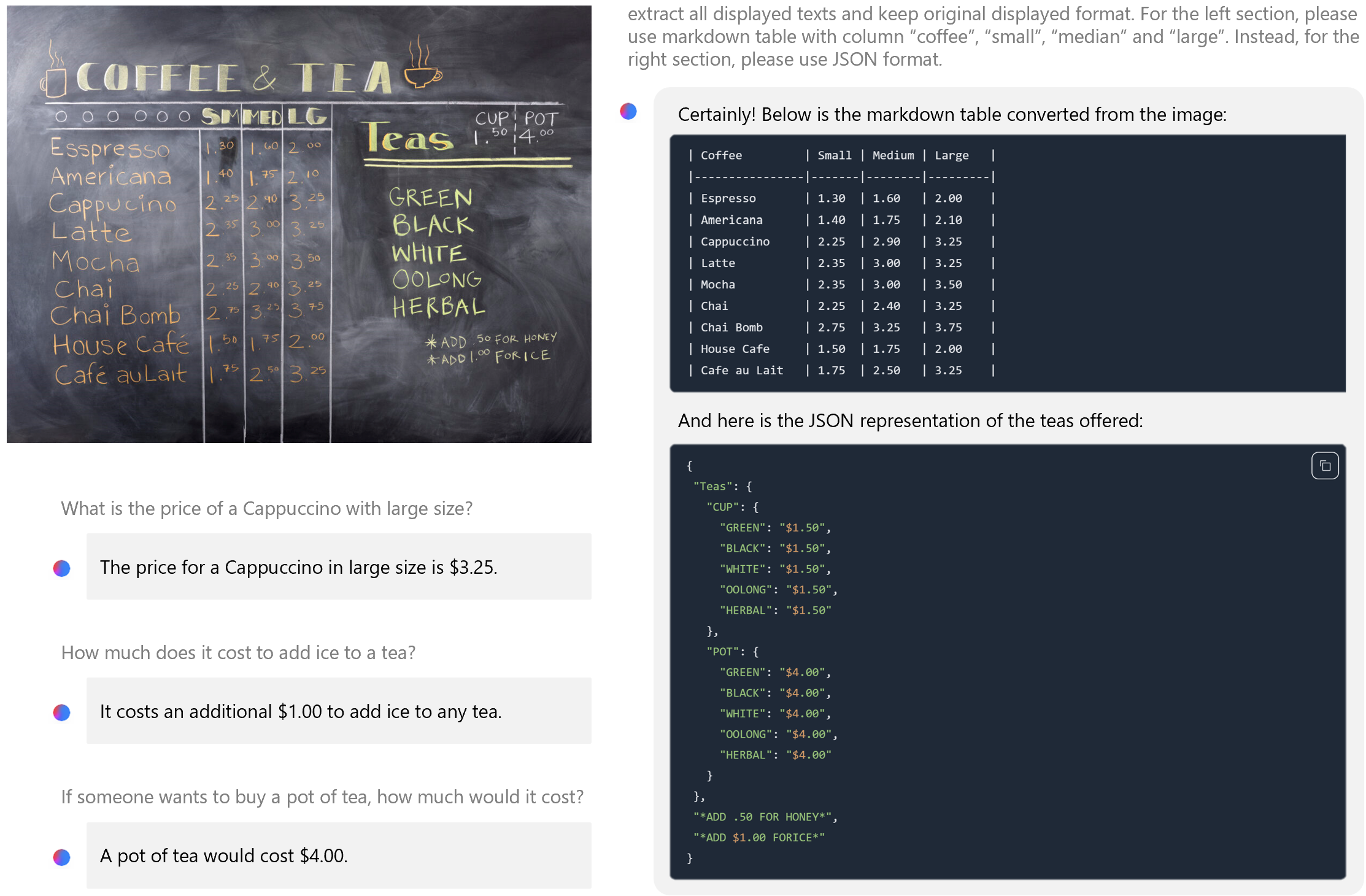}
    \caption{The demo case shows \phivision's capability in natural image understanding and reasoning.}
    \label{fig:v-safety-pt}
\end{figure}

\subsection{Technical Specifications}

\paragraph{Architecture}

The \textbf{\phivision} (4.2B parameters) is a multimodal model designed to process an image/multi-image and a textual prompt as inputs, and subsequently generate textual outputs. This model is composed of two primary components: an image encoder, \emph{i.e.}, CLIP ViT-L/14~\cite{radford2021learning} and a transformer decoder, \emph{i.e.}, phi-3.5-mini. The visual tokens, once extracted by the image encoder, are then combined with text tokens in an interleaved way (no particular order for image and text tokens). To accommodate high-resolution images and various aspect ratios, a dynamic cropping strategy~\cite{dong2024internlm} is utilized to split the input image into a 2d array of blocks, where the tokens of the blocks are concatenated to represent the whole image.  For multi-image input, we simply concatenated tokens from each images together.

\paragraph{Pre-training} 

The \textbf{\phivision} model undergoes a pre-training phase using a diverse dataset, which consists of a combination of interleaved image-text documents (\emph{e.g.}, ~\cite{laurenccon2024obelics}), image-text pairs from FLD-5B ~\cite{xiao2023florence}, synthetic data derived from Optical Character Recognition (OCR) of PDF files, datasets for chart/table comprehension, and text-only data. The objective of predicting the next token is employed specifically on text tokens, while any loss associated with image tokens is disregarded during this phase. The pre-training process involves a total of $0.5T$ tokens that encompass both visual and text elements. During the pre-training phase, the maximum image resolution is capped at $1344 \times 1344$ as the majority of the training images are smaller than this resolution.

\paragraph{Post-training.} 

The \textbf{\phivision} model contains two post-training stages: supervised finetuning (SFT) and direct preference optimization (DPO). For SFT, we leveraged text SFT dataset, public multimodal instruct tuning datasets along with large-scale multimodal instruct tuning datasets that we built ourselves, covering diverse domains and tasks such as general natural image understanding, chart/table/diagram understanding/reasoning, PowerPoint understanding, multi-image comparison, video summarization and model safety. The multimodal SFT data has about a total of 33B tokens. For DPO we mainly use a text DPO dataset and a relatively smaller-scale multimodal DPO dataset. For these two stages, we jointly train multimodal tasks and text-only tasks so that the model can achieve multi-modal reasoning while maintaining language capabilities as much as possible. 

\subsection{Academic benchmarks}

\subsubsection{Single-image Benchmarks}
We report in Table~\ref{tab:mm-benchmarks} the evaluation results of Phi-3.5-Vision on nine open-source academic benchmarks. These benchmarks evaluate reasoning and perceptual capabilities on visual and text inputs and can be grouped in three categories: Science, Charts, and Generic knowledge. We compare Phi-3.5-Vision with the following baselines: MM1-3B-Chat~\cite{mckinzie2024mm1}, MM1-7B-Chat~\cite{mckinzie2024mm1}, Llava-1.6 Vicuna 7B~\cite{liu2023improved}, Llava-1.6 Llama3-8B~\cite{liu2024llavanext}, Qwen-VL-Chat~\cite{bai2023qwenvl}, Claude 3 Haiku~\cite{anthropic2024claude}, Gemini 1.0 Pro V~\cite{team2023gemini}, and GPT-4O. Our performance quality assessment setup used the same evaluation pipeline for all the baselines to ensure a fair comparison, with the exception of MM1-3B-Chat. We just copied and pasted their published numbers since the model is not publicly available.

Our evaluation setup aimed to mimic scenarios where regular users interact with a multi-modal model, i.e., users who are not experts in prompt engineering or know special techniques that can improve performance. For this reason, we adopted the evaluation setting used in Llava-1.5~\cite{liu2023improved}. In this setup, the prompts include instructions to select a single letter corresponding to an answer from a list of given options, or answer with a single word or phrase. In our prompts, we did not use specific tokens for multiple-choice questions. Moreover, we did not scale or pre-process any image in our benchmarking system. We placed the images as the first item in the prompts, except on the MMMU dataset where the prompts interleave the images anywhere in the question or the answers. Lastly, our evaluation setup only considered a 0-shot format. Because of these evaluation parameters, our reported numbers can differ from the published numbers of the considered baselines. As we can seen, our Phi-3.5-Vision achieves super competitive results on all benchmarks and outperform other competitor models on most benchmarks while being smaller.

\subsubsection{Multi-image Benchmarks}
We report in Table~\ref{tab:mm-multi-benchmarks} the evaluation results of Phi-3.5-Vision on one latest academic multi-image benchmark and one video benchmark. These benchmarks evaluate perceptual capabilities on multiple image/frames and text covering a wide range of general scenarios (e.g., Art and Style recognition, Forensic detection, and video understanding). We compare Phi-3.5-Vision with the following baseline methods: Llava Interleave-Qwen 7B \cite{li2024llava}, InternVL2 4B and 8B \cite{chen2024far}, Gemini 1.5 Flash \cite{team2023gemini}, GPT-4o-mini, Claude 3.5 Sonnet \cite{anthropic2024claude}, Gemini 1.5 Pro \cite{team2023gemini}, and GPT-4O. Line in the single-frame evaluation case, our performance quality assessment setup used the same evaluation pipeline for all the baselines to ensure a fair comparison.

Our evaluation setup for multi-image also followed the Llava setup where  prompts include instructions to select a single letter corresponding to an answer from a list of given options, or answer with a single word or phrase. Moreover, we did not use specific tokens for multiple-choice questions and we did not scale or pre-process any image in our benchmarking system. For most of the benchmarks, we placed the images as the first item in the prompts.

The evaluation pipelines for BLINK and VideoMME benchmarks differ from those published. In the case of BLINK, we do not use ChatGPT as the final answer selection mechanism. Instead, we instruct the evaluated model to select one answer directly from the given choices. The reason is that in this manner we ensure that the mistakes or successes come solely by the evaluated model. For the VideoMME benchmark, we extracted 16 frames from the video by sampling frames at a given rate that ensures a uniform time coverage of the entire video. We used 16 frames since this is the maximum number of images a prompt can contain for Azure OpenAI models. Unlike the proposed evaluation in VideoMME that uses the maximum number of frames a model can accept, we always pass the same amount of frames across all the considered model baselines. In this way we ensure the evaluations are fair since all the models receive the exact same input information (i.e., the prompt and set of images). As shown in Table~\ref{tab:mm-multi-benchmarks}, our Phi-3.5-Vision performs very competitively or outperforms baseline models under the similar model size in multi-image understanding scenarios as well.

\begin{table}[t]
\begin{center}
\begin{adjustbox}{width=1.0\textwidth,center}
\begin{tabular}{ c||cccccccccc } 

\label{tbl:phi-v-benchmarks}

\\[10ex]
& \rothead{\makecell{\phivision\\ \footnotesize 4.2b}} & \rothead{\makecell{MM1-3B-Chat\\ \footnotesize 3.6b~\cite{mckinzie2024mm1}}} &  
\rothead{\makecell{MM1-7B-Chat\\ \footnotesize 7.6b~\cite{mckinzie2024mm1}}} &
\rothead{\makecell{LLaVA-1.6\\ \footnotesize Vicuna-7b~\cite{liu2023improved}}} & \rothead{\makecell{LLaVA-Next \\ \footnotesize LLama3-8b~\cite{liu2024llavanext}}} &  \rothead{\makecell{Qwen-VL-Chat\\ \footnotesize 9.6b~\cite{bai2023qwenvl}}} &\rothead{\makecell{Claude 3 haiku \\ \footnotesize~\cite{anthropic2024claude}}} &\rothead{\makecell{Gemini 1.0 Pro V \\ \footnotesize ~\cite{team2023gemini}}}   &  \rothead{\makecell{GPT-4O \\ \footnotesize 2024-05-13}} \\

\hline & \\[-1.5ex]

\datasetcell{\small MMMU}{\scriptsize val}{\cite{yue2023mmmu}} & 43.0 & 33.9& 37.0& 34.2& 36.4& 39.0& 40.7& 42.0& 61.8\\
\datasetcell{\small ScienceQA}{\scriptsize test}{\cite{lu2022learn}}  & 91.3& 69.4& 72.6& 70.6& 73.7& 67.2& 72.0& 79.7& 88.5\\
\datasetcell{\small MathVista}{\scriptsize testmini}{\cite{lu2024mathvista}} & 43.9& 32.0& 35.9& 31.5& 34.8& 29.4& 33.2& 35.0& 54.4\\
\datasetcell{\small Inter-GPS}{\scriptsize test}{\cite{lu2021intergps}} & 36.3& -& -& 20.5& 24.6& 22.3& 32.1& 28.6& 46.9\\
\hline & \\[-1.5ex]

\datasetcell{\small MMBench}{\scriptsize dev-en}{\cite{liu2024mmbench}} & 81.9& 75.9& 79.0& 76.3& 79.4& 75.8& 62.4& 80.0& 88.4 \\
\datasetcell{\small POPE}{\scriptsize test}{\cite{li2023evaluating}} & 86.1& 87.4& 86.6& 87.2& 87.0& 82.6& 74.4& 84.2& 87.0\\
\hline & \\[-1.5ex]

\datasetcell{\small AI2D}{\scriptsize test}{\cite{kembhavi2016diagram}} & 78.1& -& -& 63.1& 66.9& 59.8& 60.3& 62.8& 82.8\\
\datasetcell{\small ChartQA}{\scriptsize test}{\cite{masry-etal-2022-chartqa}} & 81.8& -& -& 55.0& 65.8& 50.9& 59.3& 58.0& 64.0\\
\datasetcell{\small TextVQA}{\scriptsize test}{\cite{singh2019vqa}} & 72.0& 71.9& 72.8& 64.6& 55.7& 59.4& 62.7& 64.7& 75.6\\

\end{tabular}
\end{adjustbox}
\end{center}
\caption{Comparison results on public MLLM benchmarks. All the reported numbers are produced with the exact same pipeline to ensure that the numbers are comparable except for MM1-3B-Chat~\cite{mckinzie2024mm1} and MM1-7B-Chat~\cite{mckinzie2024mm1}, which are not publicly available. We adopted the evaluation setting used in Llava-1.5~\cite{liu2023improved}, without any specific prompt or pre-processing image for all results. These numbers might differ from other published numbers due to slightly different prompts.}
\label{tab:mm-benchmarks}
\end{table}

\begin{table}[t]
\begin{center}
\begin{adjustbox}{width=1.0\textwidth,center}
\begin{tabular}{ c||ccccccccc } 

\label{tbl:phi-multi-benchmarks}

\\[10ex]
& \rothead{\makecell{\phivision\\ \footnotesize 4.2b}} & \rothead{\makecell{Llava-interleave\\ \footnotesize Qwen 7b~\cite{li2024llava}}} &  
\rothead{\makecell{InternVL2\\ \footnotesize 4b~\cite{chen2024far}}} &
\rothead{\makecell{InternVL2\\ \footnotesize 8b~\cite{chen2024far}}} & \rothead{\makecell{Gemini 1.5 \\ \footnotesize Flash~\cite{team2023gemini}}} &  \rothead{\makecell{GPT4O mini\\ \footnotesize 2024-07-18}} &\rothead{\makecell{Claude 3.5 \\ \footnotesize Sonnet ~\cite{anthropic2024claude}}} &\rothead{\makecell{Gemini 1.5 Pro  \\ \footnotesize ~\cite{team2023gemini}}}  &  \rothead{\makecell{GPT-4O \\ \footnotesize 2024-05-13}} \\

\hline & \\[-1.5ex]
\datasetcell{\small BLINK}{\scriptsize val}{\cite{fu2024blink}} & 57.0 & 53.1 & 45.9  & 45.4 & 45.8  & 51.9 & 56.5 & 61.0 & 63.2\\
% \datasetcell{\small SeedBench2}{\scriptsize test}{\cite{li2024seed}}  & 57.9  & 65.1 & 65.6 & 67& 68.8& 66.1 & 57.6& &75.9\\
\datasetcell{\small VideoMME}{\scriptsize test}{\cite{fu2024video}} & 50.8& 50.2 & 49.9&52.6 & 62.3& 61.2& 55.9 & 62.6 & 68.4\\

\end{tabular}
\end{adjustbox}
\end{center}
\caption{Comparison results on public multi-image/video MLLM benchmarks. All the reported numbers are produced with the exact same pipeline to ensure that the numbers are comparable.}
\label{tab:mm-multi-benchmarks}
\end{table}

\subsection{Safety}
To ensure the integration of \textbf{\phivision} aligns with Microsoft's Responsible AI (RAI) principles, we involved safety post-training in both Supervised Fine-Tuning (SFT) stage and Direct Preference Optimization (DPO) stage. In creating the safety training datasets, we utilized not only the text-only RAI datasets, but also a variety of in-house Multi-Modal (MM) RAI datasets that cover various harm categories identified in both public and internal MM RAI benchmarks. For the purpose of RAI evaluation, we performed a rigorous quantitative assessment on both public and internal benchmarks, this was done in conjunction with a human evaluation conducted by Microsoft's internal red team.

\begin{table}
\begin{center}
    \begin{adjustbox}{width=1.0\textwidth,center}
    \setlength\extrarowheight{6pt}
        \begin{tabular}{ c||ccccc } 
        & \makecell{\phivision \\ \footnotesize 3.8b+0.3b}& \makecell{\phivision~w/o safety \\ \footnotesize 3.8b+0.3b} & \makecell{Llava-1.6 Vicuna \\ \footnotesize 7b+0.3b } & \makecell{Qwen-VL-Chat\\ \footnotesize 7.7b+1.9b } & \makecell{GPT4-V \\ \footnotesize N/A}  \\
        \hline & \\[-3.5ex]
        Internal (private) &8.16 & 7.06  & 5.44 & 7.27 &  8.55  \\
        RTVLM (public) &5.44 & 3.56  &  3.86& 4.78 & 6.81  \\
        VLGuard (public) &9.10 & 4.66 & 5.62 & 8.33  & 8.90   \\
        \end{tabular}
    \end{adjustbox}
\end{center}
\caption{Comparison results on public and private multi-modal RAI benchmarks. Note that all metrics in the table are [0,10] and a higher value indicates a better performance.}
\label{tab:mmrai-benchmarks}
\end{table}

In Table \ref{tab:mmrai-benchmarks}, we present the evaluation outcomes of \phivision on three MM RAI benchmarks: one internal and two public benchmarks (specifically, RTVLM \cite{li2024red} and VLGuard \cite{zong2024safety}). We juxtapose these results with those of other open-source models such as Llava-1.5 \cite{liu2023improved}, Llava-1.6 \cite{liu2024llavanext}, Qwen-VL-Chat \cite{bai2023qwenvl}, and GPT4-V\cite{gpt4v}. The results clearly indicate that safety post-training notably enhances the RAI performance of \phivision across all RAI benchmarks. In Figure \ref{fig:v-safety-pt}, we further breakdown the performance across different RAI categories of the VLGuard and Internal benchmarks, demonstrating that safety post-training can aid \phivision in improving RAI performance in nearly all categories.
 
\begin{figure}[h]
    \centering
    \includegraphics[width=0.98\textwidth]{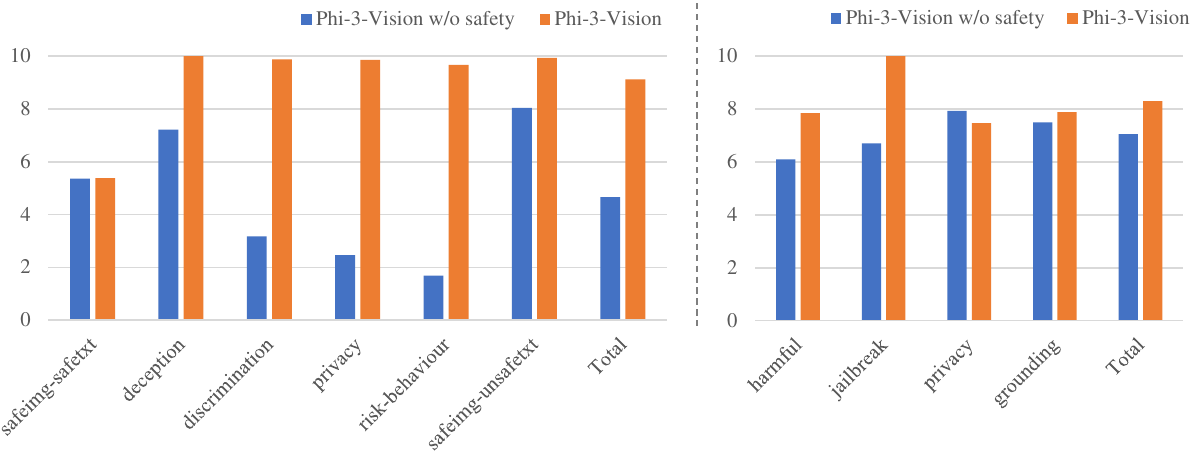}
    \caption{Comparison of categorized RAI performance of \phivision with and without the safety post-training on the VLGuard (left) and Internal (right) benchmark, respectively.  It clearly indicates that safety post-training can enhance the RAI performance across nearly all the RAI categories.}
    \label{fig:v-safety-pt}
\end{figure}

\subsection{Weakness}
Regarding the multi-modal LLM capabilities of our \phivision, it performs admirably across various fields. However, we have identified certain limitations, particularly with questions necessitating high-level reasoning abilities. Additionally, the model has been observed to occasionally generate ungrounded outputs, making it potentially unreliable in sensitive areas, such as finance. To mitigate these issues, we will incorporate more reasoning-focused and hallucination-related DPO data into post-training in the future. 

From a responsible AI standpoint, whilst safety post-training has made significant strides, our \phivision occasionally fails to refrain from answering harmful or sensitive inquiries. Examples of such occasions include deciphering particular types of captcha and describing scam images containing disinformation or hallucination. We find that this issue partly arises from the capabilities, such as OCR, acquired during the training process with normal instruct tuning datasets, which can be regarded as the trade-off between helpfulness and harmlessness. Moving forward, we need to further explore this area to achieve a better balance. 

\bibliographystyle{alpha}
\bibliography{mainbib}

\appendix
\section{Example prompt for benchmarks} \label{sec:prompt}
\begin{AIbox}{}
\tt \footnotesize 
Question:
 
Solve for $x$: $(-\frac{1}{3})(-4 -3x)=\frac{1}{2}$
 
Options:
 
A. $-\frac{5}{6}$

B. $\frac{7}{6}$

C. $\frac{5}{3}$

D. $\frac{1}{6}$
 
Answer: A

Question:

Which of the following is the body cavity that contains the pituitary gland?

Options: 

A. Abdominal

B. Cranial

C. Pleural

D. Spinal
 
Answer: B

Question:
 
Where was the most famous site of the mystery cults in Greece?

Options: 
 
A. Ephesus

B. Corinth

C. Athens

D. Eleusis

Answer:
 
\end{AIbox}

\section{Authors (alphabetical)}

\begin{tabular}{>{\raggedright\arraybackslash}p{5cm} 
                 >{\raggedright\arraybackslash}p{5cm} 
                 >{\raggedright\arraybackslash}p{5cm}}
Marah Abdin & Xin Jin & Adil Salim \\
Jyoti Aneja & Nikos Karampatziakis & Michael Santacroce \\
Hany Awadalla & Piero Kauffmann & Shital Shah \\
Ahmed Awadallah & Mahoud Khademi & Ning Shang \\
Ammar Ahmad Awan & Dongwoo Kim & Hiteshi Sharma \\
Nguyen Bach & Young Jin Kim & Yelong Shen \\
Amit Bahree & Lev Kurilenko & Swadheen Shukla \\
Arash Bakhtiari & James R. Lee & Xia Song \\
Jianmin Bao & Yin Tat Lee & Masahiro Tanaka \\
Harkirat Behl & Yuanzhi Li & Andrea Tupini \\
Alon Benhaim & Yunsheng Li & Praneetha Vaddamanu \\
Misha Bilenko & Chen Liang & Chunyu Wang \\
Johan Bjorck & Lars Liden & Guanhua Wang \\
S\'ebastien Bubeck & Xihui Lin & Lijuan Wang \\
Martin Cai & Zeqi Lin & Shuohang Wang \\
Qin Cai & Ce Liu & Xin Wang \\
Vishrav Chaudhary & Liyuan Liu & Yu Wang \\
Dong Chen & Mengchen Liu & Rachel Ward \\
Dongdong Chen & Weishung Liu & Wen Wen \\
Weizhu Chen & Xiaodong Liu & Philipp Witte \\
Yen-Chun Chen & Chong Luo & Haiping Wu \\
Yi-Ling Chen & Piyush Madan & Xiaoxia Wu \\
Hao Cheng & Ali Mahmoudzadeh & Michael Wyatt \\
Parul Chopra & David Majercak & Bin Xiao \\
Xiyang Dai & Matt Mazzola & Can Xu \\
Matthew Dixon & Caio C\'esar Teodoro Mendes & Jiahang Xu \\
Ronen Eldan & Arindam Mitra & Weijian Xu \\
Victor Fragoso & Hardik Modi & Jilong Xue \\
Jianfeng Gao & Anh Nguyen & Sonali Yadav \\
Mei Gao & Brandon Norick & Fan Yang \\
Min Gao & Barun Patra & Jianwei Yang \\
Amit Garg & Daniel Perez-Becker & Yifan Yang \\
Allie Del Giorno & Thomas Portet & Ziyi Yang \\
Abhishek Goswami & Reid Pryzant & Donghan Yu \\
Suriya Gunasekar & Heyang Qin & Lu Yuan \\
Emman Haider & Marko Radmilac & Chenruidong Zhang \\
Junheng Hao & Liliang Ren & Cyril Zhang \\
Russell J. Hewett & Gustavo de Rosa & Jianwen Zhang \\
Wenxiang Hu & Corby Rosset & Li Lyna Zhang \\
Jamie Huynh & Sambudha Roy & Yi Zhang \\
Dan Iter & Olatunji Ruwase & Yue Zhang \\
Sam Ade Jacobs & Olli Saarikivi & Yunan Zhang \\
Mojan Javaheripi & Amin Saied & Xiren Zhou \\

\end{tabular}

\section{Acknowledgements}
We would like to thank Zhuohan Li, Simon Mo from UC Berkeley and Kaichao You from Tsinghua University for sharing their insights on the vLLM kernel.

\end{document}